\documentclass[10pt,journal,compsoc]{IEEEtran}
\ifCLASSOPTIONcompsoc
  \usepackage[nocompress]{cite}
\else
  \usepackage{cite}
\fi
\ifCLASSINFOpdf
\else
\fi
\usepackage{multirow,palatino, epsfig, latexsym, algorithmic, algorithm , epstopdf, amsmath, amssymb, color, amsthm,graphicx}

\newcommand{\ignore}[1]{}

\newtheorem{myRemark}{Remark}

\newtheorem{myDefinition}{Definition}

\begin{document}
\title{Optimization of distributions differences for classification}

\author{Mohammad Reza Bonyadi, Quang M. Tieng, David C. Reutens,

%\IEEEcompsocitemizethanks{\IEEEcompsocthanksitem }% <-this % stops an unwanted space 
\thanks{All authors are with the Centre for Advanced Imaging (CAI), the University of Queensland, Brisbane, QLD 4072, Australia. M. R. Bonyadi (reza@cai.uq.edu.au, rezabny@gmail.com) is also with the Optimisation and Logistics Group, The University of Adelaide, Adelaide 5005, Australia. }}

%\markboth{Journal of \LaTeX\ Class Files,~Vol.~13, No.~9, September~2014}%
%{Shell \MakeLowercase{\textit{et al.}}: Bare Demo of IEEEtran.cls for Computer Society Journals}

\IEEEtitleabstractindextext{%
\begin{abstract}
In this paper we introduce a new classification algorithm called Optimization of Distributions Differences (ODD). The algorithm aims to find a transformation from the feature space to a new space where the instances in the same class are as close as possible to one another while the gravity centers of these classes are as far as possible from one another. This aim is formulated as a multiobjective optimization problem that is solved by a hybrid of an evolutionary strategy and the Quasi-Newton method. The choice of the transformation function is flexible and could be any continuous space function. We experiment with a linear and a non-linear transformation in this paper. We show that the algorithm can outperform 6 other state-of-the-art classification methods, namely naive Bayes, support vector machines, linear discriminant analysis, multi-layer perceptrons, decision trees, and k-nearest neighbors, in 12 standard classification datasets. Our results show that the method is less sensitive to the imbalanced number of instances comparing to these methods. We also show that ODD maintains its performance better than other classification methods in these datasets, hence, offers a better generalization ability. 
\end{abstract}

%\begin{IEEEkeywords}
%Supervised learning, neural networks, optimisation, evolutionary strategy.
%\end{IEEEkeywords}
}

% make the title area
\maketitle
\IEEEdisplaynontitleabstractindextext
\IEEEpeerreviewmaketitle

%\IEEEraisesectionheading{
\section{Introduction}
\label{sec:intro}
%}

\IEEEPARstart{T}{he} ultimate goal of a supervised classification method is to identify to which class a given instance belongs based on a given set of correctly labeled instances. A classifier in this paper is defined \cite{ng2002discriminative} as follows:
\ignore{
While some methods (known as discriminative classification methods) generate the exact label for each instance, some others (known as generative methods) provide $ p(y|x) $, $ y $ is the label for a given instance $ x $, it is up to the user to interpret this probability to generate the final label \cite{ng2002discriminative}. 
\begin{myDefinition}
(Generative classifier) 
Let $ S,S_{1},S_{2},...,S_{c} $ be sets of instances, $ |S_i|=m_i $, $ S_{i} \cap S_{j} = \emptyset $ for all $ i,j \in \{1,...,c\} $, $ i \ne j $, $ \cup_{i=1}^{c}S_i=S $. A classifier $C_g$ aims to guarantee
\begin{equation*}
\forall i \in \{1,...,c\}, \forall x \in S_i, P(i|x) \ge P(j|x).
\end{equation*}	
\end{myDefinition}
}
\begin{myDefinition}
	(Discriminative classifier)
Let $ S,S_{1},S_{2},...,S_{c} $ be sets of instances, $ |S_i|=m_i $, $ S_{i} \cap S_{j} = \emptyset $ for all $ i,j \in \{1,...,c\} $, $ i \ne j $, $ \cup_{i=1}^{c}S_i=S $. A classifier $\psi_\kappa(x)$, $\psi_\kappa:S \to \mathbb{R}$, aims to guarantee
\begin{equation*}
\forall i \in \{1,...,c\}, \forall x \in S_i, P(\psi_\kappa(x)=i|x)=1.
\end{equation*}	where $\kappa$ is a set of configurations for the procedure $\psi_\kappa(x)$, and $P$ is the probability measure.
\end{myDefinition}

 In this paper, we consider a special case of classification problems where all members of $ S $ are in $ \mathbb{R}^n $ (so called feature space). We also assume that feasible values for $ x^i $ (called a variable throughout this paper), the $ i^{th} $ element of the instance $ \vec{x} $, are ordered by the operator "$ \le $" (i.e., $ x^i $ is not categorical). 

In reality, only a subset of $ S $ is given (the training set) for which the classes are known. It is then critical to find the best $\kappa$ in a way that $\psi_\kappa(\vec{x})$ is the true class of any $ \vec{x} $ in the training set and, ideally, all possible instances in $ S $. However, the distribution of instances in each class $ S_i $ is unknown, making the best estimation of $ \kappa $ challenging. 

Classification is required in many real-world problems. Although there have been many classification methods proposed to date, such as multi-layer perceptrons \cite{haykin2004comprehensive}, decision trees, support vector machines \cite{suykens1999least}, and extreme learning machines \cite{huang2006extreme}, there are still some limitations associated with many of these methods. Some methods are sensitive to imbalanced number of instances in each class \cite{branco2016survey}. Also, non-linear classification methods outperform linear classification methods in the training set, however, may end up having worse performance when they are applied to new instances, an issue known as overfitting \cite{overfitHawkins}. 

We propose a new classification algorithm called optimization of distributions differences (ODD) in this paper. ODD aims to optimize distribution of instances in different classes to ensure they do not overlap. ODD finds a transformation, $ F:\mathbb{R}^n \to \mathbb{R}^p $, $ n $ is the number of dimensions (features) for each instance and $ p $ is a positive integer, in a way that the distance between the gravity centers of the instances in different classes is maximized while the spread of the instances within the same class is minimized. If such transformation exists, the instances could be assigned to each class based on their distances to the centers of the classes. We formulate the optimization of this transformation as a multiobjective optimization problem with two sets of objectives. The first set of objectives ensure that the centers of different classes are as far as possible from one another by defining $ c(c-1)/2 $ objectives, $c$ is the number of classes. The second set of objectives ensure that the spread of instances within each class is minimized, defined by the norm of the eigenvalues of the covariance matrix of instances in each class, that adds $ c $ extra objectives to the system. We solve this problem using a combination of evolutionary algorithms \cite{beyer2002evolution} and conjugate gradient methods \cite{fletcher2013practical}. We experiment with linear and non-linear transformations and show that the method can outperform existing classification methods when they are applied to 12 standard classification benchmark problems and 4 artificial classification problems. We also show that the algorithm is not sensitive to the imbalance number of instances in the classes, assuming that the given instances for training represent the classes distribution parameters to some extent. Our experiments indicate that the method outperforms both non-linear and linear classifiers in terms of generalization ability.

We structure the paper as follows: section \ref{sec:background} outlines some background information on classification methods and optimization algorithms we will use in this paper. Section \ref{sec:proposed} provides details of our proposed method, including the model, definitions, and optimization. Section \ref{sec:experiment} reports and discusses comparative results among the multiple classification methods on 12 standard benchmark classification problems. Sensitivity to imbalance number of instances in each class and overfitting are also discussed in that section. Section \ref{sec:conclusions} concludes the paper and discusses potential future directions.

\section{Background}
\label{sec:background}
This section provides some background information about existing classification and optimization methods.
\subsection{Classification methods}
In this section, we describe classification methods that we use for comparison purposes.
\subsubsection{K-nearest neighbor (Knn)}
K-nearest neighbor (Knn) \cite{altman1992introduction} works based on the assumption that instances of each class are surrounded mostly by the instances from the same class. Hence, given a set of training instances in the feature space and a scalar $ k $, a given unlabeled instance is classified by assigning the label which is most frequent among the $ k $ training samples nearest to that instance. Among many different measures used for distance between instances, Euclidean distance is the most frequently used for this purpose.
\subsubsection{Naive Bayes (NBY)}
NBY is a classification algorithm that works based on the Bayes theorem\cite{hand2001idiot}. The aim of the algorithm is to find the probability that a give instance $\vec{x}$ belongs to a class $c$, i.e., $P(c|\vec{x})$. To calculate this, NBY uses the Bayes theorem as $P(c|\vec{x})=\frac{P(\vec{x}|c)P(c)}{P(\vec{x})}$. The value of $P(c)$, $P(\vec{x})$, and $P(\vec{x}|c)$ can be all estimated from the given instances in the training set. As the instance $\vec{x}$ is in fact a vector that contains multiple variables, $P(\vec{x}|c)$ is estimated by $P(x_1|c)\times P(x_2|c) \times ... \times P(x_n|c)$ that indeed ignores the dependency among variables, a "naive" assumption. 

\subsubsection{Support vector machines (SVM)}
The aim of SVM \cite{suykens1999least} is to find a hyperplane defined by the normal vector $ \vec{\omega} $ that could separate two class of instances \cite{suykens1999least}. The separation is determined by the sign of $ \vec{\omega} \vec{x}^T + r $ that indicates to which side of the hyperplane the instance $ \vec{x} $ belongs. In other words, given a set of instances and their classes (supervised learning), the algorithm outputs an optimal hyperplane which categorizes instances. 

One way to extend this algorithm to deal with multiclass classification problems is to use one-vs-all or one-vs-one strategies proposed in \cite{multiclassSVMComparisons}. 

\subsubsection{Multi-layer percenptrons (MLP)}
MLP aims to optimize the parameters of a mapping from a set of input instances to their provided outputs to estimate the Bayes optimal discriminant \cite{ruck1990multilayer}. The mapping can be linear or non-linear and could be presented in multiple layers. The algorithm minimizes the mean square of error between the generated outputs and expected outputs for each instance. One of the most frequently used optimization methods in MLPs is the Levenberg-–Marquardt \cite{levenberg1944method,marquardt1963algorithm}, that is also used in this paper. See \cite{pal1992multilayer} for more details. 

\subsubsection{Linear discriminant analysis (LDA)}
The aim of LDA is to calculate $ \vec{\omega} $ for which $ \vec{\omega}\vec{x}^T>k $ if the instance $ \vec{x} $ belongs to the second class. Assuming that the conditional probability $ p({\vec {x}}|y=0) $ and $ p({\vec {x}}|y=1) $ ($y$ is the label of $\vec{x}$) are both normally distributed with mean and covariance parameters $ \left({\vec {\mu }}_{1},\Sigma_1\right) $ and $ \left({\vec {\mu }}_{2},\Sigma_2\right) $, Fisher \cite{fisher1936use} proved that $ \vec{\omega}=(\Sigma_1+\Sigma_2)^{-1}(\vec{\mu}_2-\vec{\mu}_1) $ and $ k={\frac {1}{2}}{\vec {\mu }}_{2}^{T}\Sigma_2^{-1}{\vec {\mu }}_{2}-{\frac {1}{2}}{\vec {\mu }}_{1}^{T}\Sigma_1^{-1}{\vec {\mu }}_{1} $ could distinguish between the two classes. Considering $S_W=(\Sigma_1+\Sigma_2)$ as a measure for the within-class spread and $S_B=(\vec{\mu}_2-\vec{\mu}_1)(\vec{\mu}_2-\vec{\mu}_1)^T$ ($T$ is the transpose operator) as a measure for between-class spread, Fisher's value for $\vec{\omega}$ ensures that $\frac{W^TS_BW}{W^TS_WW}$ is maximized. $ \vec{\omega} $ is the norm of a hyperplane that discriminates the two classes and $ k $ is the shift to ensure this hyperplane is between the two classes. 

If $ \Sigma_1 $ and $ \Sigma_2 $ are small then $ (\Sigma_1+\Sigma_2)^{-1} $ becomes singular that lead to vanishing the impact of $ (\vec{\mu}_2-\vec{\mu}_1) $, i.e., a solution that leads to singular $ (\Sigma_1+\Sigma_2)^{-1} $ dominates all other solutions, no matter the distance between the centers of the classes \cite{gao2006direct}. This, however, is not desirable as it is important that the classes centers are apart from one another to be able to distinguish between them. This scenario occurs particularly when the number of instances in a class is smaller than the number of dimensions $n$. Also, the threshold $ k $ is effective only if the distribution of the classes are similar, that might not be the case in many datasets.

One way to extend this algorithm to deal with multiclass classification problems is to use one-vs-all or one-vs-one strategies proposed in \cite{multiclassSVMComparisons}. Direct-LDA is another version of LDA that can handle multiple classes.

\subsubsection{Direct-LDA}
Direct-LDA is a variant of LDA that handles multiple classes \cite{yu2001direct}. Between-class spread for Direct-LDA is formulated by:
\begin{equation}
S_B=\frac{1}{c}\sum_{i=1}^{c}(\mu_i-\mu)(\mu_i-\mu)^T
\end{equation}
where $ \mu_i $ is the average of the instances in the class $ i $, $ \mu $ is the average of all $ \mu_i $s. The within-class spread is defined by:
\begin{equation}
S_W=\sum_{i=1}^{c}Cov(X^i)
\end{equation}
where $ X^i $ is a $m_i \times n$ matrix, $m_i$ is the number of instances in the class $i$, each row is an instance of the class $i$, and $Cov(.)$ is the covariance operator. In the multiclass case, the optimum value for $ \omega $ (that is not a vector anymore but a $ n \times (c-1) $ matrix) is then the first $ c-1 $ eigenvectors corresponding to the $ c-1 $ largest eigenvalues of $S_W^{-1}S_B$. If the number of dimensions is smaller than the number of classes then the algorithm might not find effective $\omega$ to distinguish between classes \cite{gao2006direct}. Another limitation with this formulation is that, if the number of dimensions is smaller than the number of instances in one of the classes, the covariance matrix for that class becomes non-full rank. In addition, if the distance between two classes is large it may dominate the spread of the classes ($S_B$) and lead to ineffective transformation \cite{li2006using}.

\ignore{		
\begin{itemize}
	\item if covariance matrices become very small, the impact of the between-class objective vanishes. Hence, a solution that ensures covariance of all classes is zero cannot be dominated by any other solution. However, such solution might not be optimal as it may also minimize the distance between centers, that is not desirable.
	\item the variance of between classes means could mislead the calculations. The variance is large if the classes are far from one another OR only one class is too far and the rest are close to one another. However, it is beneficial to ensure that the distance between the centers of classes is large rather than the variance of these means is large. 
	\item Multi class only supports similar covariances. 
	\item Fisher assumed similar covariances and then defined the the threshold, that is problematic 
	\item number of dimensions larger than number of instance in one class then the covariance becomes non-full rank
\end{itemize}
}

\subsubsection{Decision tree (DTR)}
Decision tree (DTR) is a tree structure for which each interior node corresponds to one of the input variables and each leaf represents a class label. The outgoing edges from each interior node represent the decision made for variable values in that node. For a given instance, a path from the root of the tree that follows the values of each variable lead to the class label for that instance. The tree is trained (e.g., by the method proposed in \cite{quinlan1986induction}) according to the given instances in the training set.
\subsection{Derivative-free optimization methods}
In this section we provide a brief background information about optimization algorithms we use in this paper.
\subsubsection{Quasi-Newton (QN)}
The aim of QN is to find a point in a search space that its gradient is zero. The method assumes that the objective function can be estimated by a quadratic function around the local optimum and finds the root of the first derivative of the objective function by generalizing the secant method. We use Broyden–-Fletcher-–Goldfarb-–Shanno (BFGS) \cite{fletcher2013practical} in this article to constrain the solutions of the secant equation as this method is frequently used in the literature and provide acceptable practical performance. The finite difference gradient approximation is usually used for objective functions that their gradient cannot be calculated analytically.

\subsubsection{Evolutionary strategy (ES)}

Evolutionary algorithms work based on a population of candidate solutions that are evolved according to some rules until they converge to an optimum solution. Examples include particle swarm optimization \cite{bonyadi2016review} and evolutionary strategy \cite{beyer2002evolution}, each has its own specific properties that make them advantageous/disadvantageous on various types of problems. The aim of these methods is to use information coded in each individual in the population (with the size $\lambda$) and update them to find better solutions. For example, evolutionary strategy (ES) generates new individuals using a normal distribution with the mean of the current location of the individual and an adaptive variance, calculated based on the distribution of "good" solutions. Covariance matrix adaptation evolutionary strategy (CMAES) employs similar idea but updates the covariance matrix of the normal distribution (rather than the variance alone) to generate new instances that accelerates convergence to local optima. This idea takes into account non-separability of dimensions during the optimization process, hence, would be more successful when the variables are interdependent. See \cite{beyer2002evolution} for detail of these methods.

\section{Proposed algorithm}
\label{sec:proposed}
We define a $\psi_\kappa$ by a tuple $ <F,f,\Omega> $, a surjective transformation $ F $, a discriminator $ f $, and an optimization problem $ \Omega $ that aims to find the best $ F $ such that 
\begin{equation*}
\forall i \in \{1,...,c\}, \forall x \in S_i, f(F(\vec{x})) = i
\end{equation*}	
where $ f $ denotes the class index of the transformed instance $ \vec{x} $. Although only $ F $ and $ f $ are required to classify a set of instances, the optimization problem, $ \Omega $, is also very important component to ensure efficiency of the model and the discriminator. 

In this paper, we consider that $ F: \mathbb{R}^n \to \mathbb{R}^p $, and $ f:\mathbb{R}^p \to \mathbb{R} $. It is usually assumed that the discriminator function $ f $ is constant while the parameters of the transformation $ F $ are formulated into $ \Omega $ and optimized through an optimization procedure. In SVM with linear kernel, for example, the function $ F $ is defined by $ F(\vec{x})=\vec{x} \vec{\omega}^T -b $, $ \vec{x}\in \mathbb{R}^n $, $ f(F(\vec{x}))=sign(F(\vec{x})) $. MLP with no hidden layer assumes that $ F(\vec{x})=t(\vec{x} M_{n \times p} + \vec{b}) $ and $ f(F(\vec{x}))=F(\vec{x}) $, where $ t $ is the activation function (usually $ tan $ or $ log $ sigmoid), and $ \Omega $ is to minimize the average of $ (f(F(\vec{x}^i))-G(\vec{x}^i))^2 $ over all given instances ($ i $) where $ G(\vec{x}^i) $ is the class of $ \vec{x}^i $. 
\ignore{
In reality, only a subset of $ S $ is given (the training set) for which the classes are known. It is then critical to design $ F $ and $ f $ in a way that $ f(F(\vec{x})) $ is the true class of any $ x $ in this given set and, ideally, all possible instances in $ S $. However, the distribution of instances in each class $ S_i $ is unknown, making the best estimation of $ F $ and $ f $ challenging. Clearly, the less instances of each class is given, the less is known about that class and, hence, more probable to mis-model the instances in that class by $ F $ and $ f $, this of course is also dependent on the given instances as well, i.e., how accurate the given instances of each class represent all possible instances in that class. 
}

Let us assume that the instances in each class $ S_i $ are random variables that follow a distribution with specific moments. The optimization problem $ \Omega $ for the optimization of distribution difference (ODD) algorithm is to find a transformation $ F:\mathbb{R}^n \to \mathbb{R}^p $ such that the the gravity center of the transformed instances by $ F $ that are in different classes are as far as possible from one another while the distance among transformed instances that are in the same class is minimized. After optimization of $ F $, a discriminator $ f $ could be simply defined by the distance between the given instances and the centers of the classes. We formulate $ F $, $ f $, and $ \Omega $ for ODD in the remaining of this section.

\subsection{The optimization problem $ \Omega $ for ODD}
Let $ X^{(k)}_{m_k \times n} $ include all given instances of the class $ k $ (a subset of $ S_k $), each row corresponds to one instance. We transform each row of this matrix by the function $ F $ to form $ Y^{(k)}_{m_k \times p} $. We define $ \vec{a}^k $, a $ p $ dimensional vector, as the center of gravity of all $ m_k $ instances in $ Y^{(k)}_{m_k \times p} $ as follows:
\begin{equation}\label{eq:proposed-a}
\vec{a}^k=\frac{1}{m_k}\sum_{i=1}^{m_k}\vec{y}_i
\end{equation}
where $ \vec{y}_i $ is the $ i^{th} $ row of $ Y^{(k)}_{m_k \times p} $. We also define the scalar $ v^k $, as the norm of the eigen values of the covariance matrix of $ Y^{(k)}_{m_k \times p} $: 
\begin{equation}\label{eq:proposed-v}
v^k=||Eig(Cov(Y^{(k)}_{m_k \times p}))||
\end{equation}
where $ Cov(.) $ is the covariance operator and $ Eig(.) $ calculates the eigen values of its input matrix. The value of $ v^k $ indicates how the instances in the class $ k $ have been spread around their center along with their most important directions (eigen vectors). The aim of ODD is to adapt the transformation $ F $ such that $ v^k $ is minimized for all $k$ while the distances among all possible gravity centers are maximized. This could be formulated by a multiobjective optimization problem:
\begin{equation}\label{eq:proposed-obj}
\Omega=
\begin{cases}
	\text{maximize } ||\vec{a}^i-\vec{a}^j|| & \text{for all } j>i\\
	\text{minimize } v^i & \text{for all } i
\end{cases}
\end{equation}
where $ i,j\in\{1,...,c\} $. The problem contains $ c $ minimization and $ \frac{c(c-1)}{2} $ maximization objectives. We use the following remarks to convert this multiobjective problem to a single objective problem:
\begin{myRemark}
	Let us assume that $ A_i(x) $ is a function and $ A_i(x)>0 $ for all $ i $ and $ x $. A solution that minimizes $ \sum_{i}A_i(x) $ is on the true Pareto front of the multiobjective optimization problem: ''minimize $ A_i(x) $ for all $ i $''
\end{myRemark}

\begin{myRemark}
	Let us assume that $ A_i(x) $ is a function and $ A_i(x)>0 $ for all $ i $ and $ x $. A solution that maximizes $ \prod_{i}A_i(x) $ is on the true Pareto front of the optimization problem: ''maximize $ A_i(x) $ for all $ i $''
\end{myRemark}	
The proofs for both these remarks are elementary and could be done by contradiction. 

Using remark 1 and remark 2, the multiobjective optimization problem defined in Eq. \ref{eq:proposed-obj} can be transformed to a single objective optimization problem defined by:
\begin{equation}\label{eq:proposed-obj-single}
\Omega=\text{minimize } \frac{\gamma+\sum_{k=1}^{c}v^k}{\left(\prod_{i}^{c}\prod_{j=i+1}^{c}||\vec{a}^i-\vec{a}^j||\right)^\frac{1}{c(c-1)}}
\end{equation}
where $ \gamma $ is a positive constant (set to 1 in our experiments) to ensure that, among all possible solutions for which $ \sum_{k=1}^{c}v^k=0 $, the one that maximizes centers distances ($ \prod_{i}^{c}\prod_{j=i+1}^{c}||\vec{a}^i-\vec{a}^j|| $) is preferred. Because the growth rate of the denominator is factorial, we have used the regulator $ \frac{1}{c(c-1)} $ to balance the growth rate of the nominator and denominator. This ensures balancing the importance of distinct gravity centers while minimizing the spread of instances within each class. The product (rather than a simple weighted summation) enforces the optimizer to find solutions that impose scattered centers for classes. This is an extremely important point as, otherwise, the optimization algorithm may find a solution that maps some of the centers close to one another while move the remaining centers far from the others, that is not desirable. 

Note that the transformation of the multiobjective optimization (Eq. \ref{eq:proposed-obj}) to its single objective form (Eq. \ref{eq:proposed-obj-single}) is effective only if the objectives are assumed to be equally important, that is the case here. 

\subsection{The discriminator function $ f $ for ODD}
\label{sec:discriminatorf}
After solving $ \Omega $ (Eq. \ref{eq:proposed-obj-single}), we need to identify to which class a given vector $ \vec{y} $ belongs. We define the function $ f=<f_1,f_2,...,f_c> $ as follows:
\begin{equation}
f_k(\vec{y})=\frac{D_k}{\sum_{j=1}^{c}D_j}
\end{equation}
where $ D_k=||\vec{y}-\vec{a}^k|| $ is the distance between $ \vec{y} $ and the center of the class $ k $. The smaller the value of $ f_k $ is, the more likely that the vector $ \vec{y} $ belongs to the class $ k $. One can calculate $ 1-f_k(\vec{y}) $ and then normalize the results to get probabilities of $ \vec{y} \in S_k $. This leads us to the following formula for $ f_k $: 
\begin{eqnarray}\label{eq:proposed-f_kfinal}
f_k(\vec{y})=1-\frac{1-\frac{D_k}{\sum_{j=1}^{c}D_j}}{\sum_{i=1}^{c}(1-\frac{D_k}{\sum_{j=1}^{c}D_j})}\\ = \frac{1}{c-1}-\frac{D_k}{(c-1)(\sum_{i=1}^{c}D_i)} \nonumber
\end{eqnarray}
Clearly, $ f_k(\vec{y}) \in [0,1] $, that could be interpreted as a measure for the probability of $ \vec{y} \in S_k $. In order to convert the results to a categorical value (i.e., converting generative results to the discriminative results), we use the threshold found in the training set that maximizes the area under the curve (AUC) and use that threshold to discriminate the final results in test cases. This thresholding strategy is used for all generative methods in this article (Direct-LDA, ODD, and MLP) unless specified.

\subsection{The transformation function $ F $ for ODD}
We consider the linear case for $F$ in this paper where $ F(\vec{x})=\vec{x}\times M_{n\times p} + \vec{r}$, $ \vec{x} $ is a $ n $ dimensional vector. Accordingly, $ \Omega $ in Eq. \ref{eq:proposed-obj-single} is dependent only on $ M_{n\times p} $ and $\vec{r} $ that introduces $ p(n+1) $ variables. In the rest of this paper, we combine $ M_{n\times p} $ and $ \vec{r} $ and denote $ M_{n' \times p} $ where $ n'=n+1 $. This, of course, would assume that an instance $ \vec{x}^i $ is presented as $ <x^i_1,...,x^i_n,1> $. One can see ODD with this setting is very similar to MLP with no hidden layer that uses a different energy function: MLP uses the mean square error of the outputs of instances and expected classes independently while ODD uses the idea of centrality of instances that are in the same class. 

Not all problems could be effectively transformed by a linear function. Hence, one may add nonlinear flexibility to $ F $ by introducing a function $ g:\mathbb{R}^p \to \mathbb{R}^p $ where $ F(\vec{x})=g(\vec{x}\times M_{n'\times p}) $. If the function $ g $ is nonlinear then the final model could classify instances that are nonlinearly separable. The choice of the function $ g $ is problem dependent and could be done through a trial and error procedure. We will test $ g(\vec{x})=tanh(\vec{x}) $ in the experiment section as a candidate to introduce non-linearity to our algorithm. This function has been frequently used for this purpose in MLP articles. Note also that, unlike other classification methods that require specific functions for their kernel, the transformation $ F $ for ODD could be defined in more generic form as the optimization function $ \Omega $ for ODD is solved by a derivative-free optimization method.

\subsection{Candidate optimization methods for ODD}
The optimization problem, $ \Omega $, introduced in Eq. \ref{eq:proposed-obj-single} is nonlinear. It is also difficult to calculate the gradient and Hessian for this equation, that leaves us with the methods that are either derivative-free or approximate gradient and Hessian. In addition, it is not clear if this optimization problem is unimodal or multimodal, make the solution even more challenging. In this paper, we use three methods to solve this optimization problem: ES, CMAES, and QN. We are not using PSO as it does not take into account the dependency among variables \cite{bonyadi2016review} that is required for the purposes of this paper. The first two methods are stochastic and population-based and have a better exploration ability than the last. However, QN could converge to a local optimum faster than other methods \cite{loshchilov2015lm}. In this section we compare the computational complexity of these methods. 

\begin{figure}
	\centering
	\begin{tabular}{c}
		\includegraphics[width=0.49\textwidth]{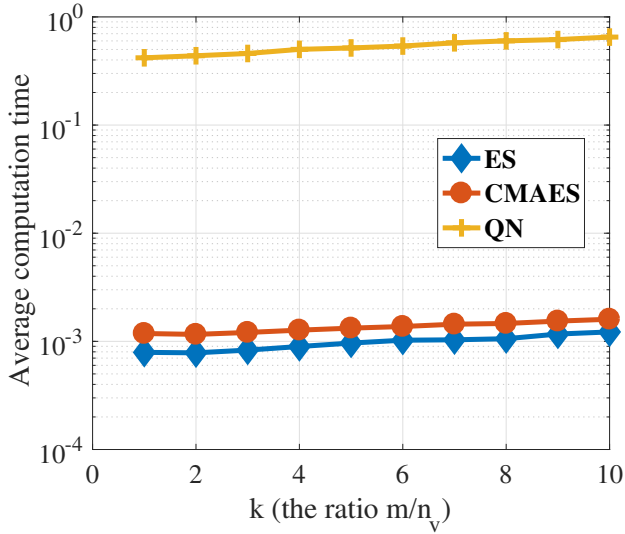}\\
		(a)\\
		\includegraphics[width=0.49\textwidth]{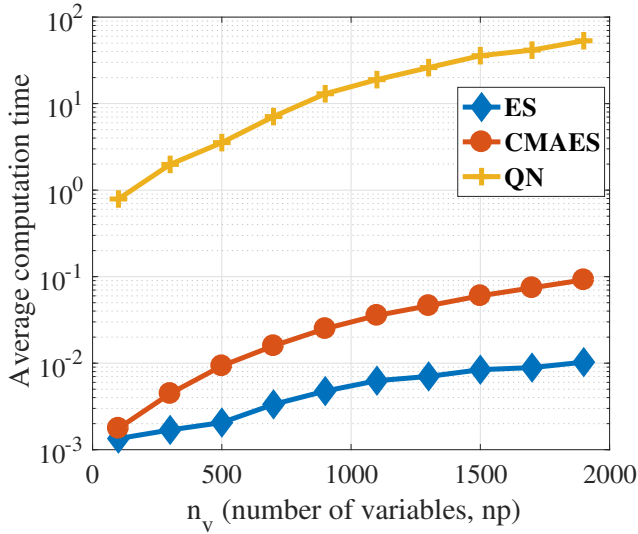}\\
		(b)
	\end{tabular}

	\caption{The average computation time (in milliseconds) for different methods at each iteration: (a) when $k$ is changed (that changes the number of instances), (b) when $n_v$ is changed.}
	\label{fig:proposed-compute-time-m}
\end{figure}

Let $ F(\vec{x})=\vec{x}M_{n \times p} $, $ n_v=np $ the number of variables to optimize, $ m=kn_v $, where $ k $ is a constant. In order to demonstrate the time complexities in practice, we designed a random dataset in which $ m=knp $ instances were uniformly randomly sampled in an $ n $ dimensional space ($ n $ dimensional uniformly sampled instances) and assigned to two classes randomly. We set $ p=2 $ in all examples and applied ES, CMAES, and QN to the objective function of ODD, each method for 5 iterations, 50 times. The function $ F $ was set as specified earlier (linear function), and $ \lambda $ was set to 50 for ES and CMAES. Figure \ref{fig:proposed-compute-time-m}(a) shows the results when $ k $ was changed from $ 1 $ to $ 10 $ and $ n=100 $. The figure shows that the average computation time for all methods is linear w.r.t $ k $. However, the calculations included much larger constant multipliers for QN that makes the algorithm significantly slower than other methods. Figure \ref{fig:proposed-compute-time-m}(b) shows the results when $ n_v $ was changed from $ 100 $ to $ 1000 $ and $ m=2000 $. The figure shows that the required time for ES is less than other methods. 

Another important factor that contributes to the performance of the optimization method is the convexity of the search space. As there is no reason to assume that the optimization problem defined in Eq. \ref{eq:proposed-obj-single} is convex, the exploration ability of the algorithm becomes important. While methods like QN are very efficient in convex spaces, they have difficulty in finding good solutions in non-convex problems. In contrast, ES and CMAES have a better exploration ability that enables them to offer better final solutions in multimodal optimization problems. It is hence beneficial to use a hybrid of ES and CMAES with QN to ensure effective exploration at the beginning of the search while better exploitation at the later stages of the search. Hence, in all of our implementations, we used CMAES in combination with QN for small problems ($ n_v \le 300 $) and ES alone for large problems ($ n_v>300 $), set experimentally. Although the number of iterations for these methods could be set according to $ n_v $, our experiments showed that 100 for CMAES, 100 for QN, and 500 for ES work efficiently for our test cases. Also, ES and CMAES are terminated if their performance is not improved by at least 0.001 in the last 20 iterations. For QN, the algorithm was terminated if the gradient value was smaller than 1e-8.
\subsection{An example}
\begin{figure}
	\centering
	\includegraphics[width=0.49\textwidth]{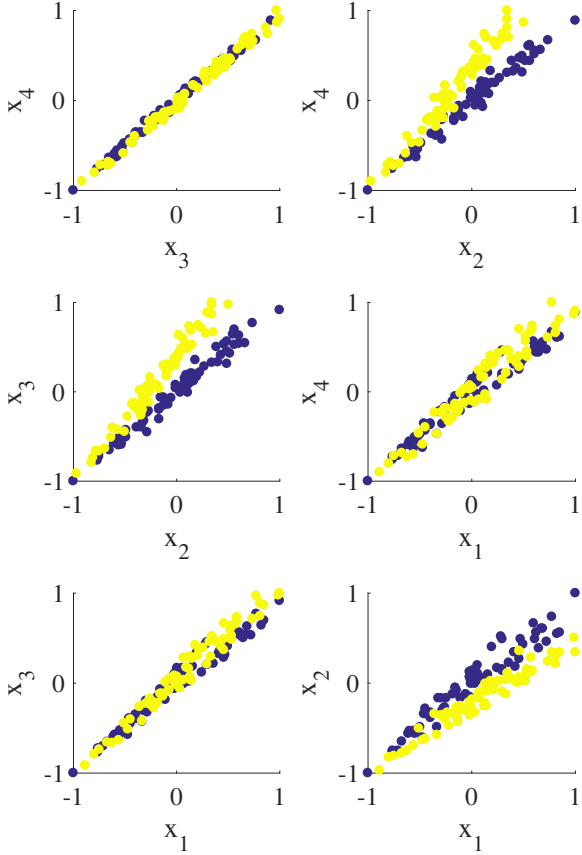} 
	\caption{Illustration of the instances from the Crab gender in four dimensions.}
	\label{fig:proposed-rawDb-Example}
\end{figure}

Let us give an example to clarify how ODD works. Assume the dataset presented in Fig. \ref{fig:proposed-rawDb-Example}\footnote{This dataset is in fact a part of the crab gender dataset \cite{allMachineLearnings} where only 4 of the attributes were used for this example. This dataset has been used in its complete form in our experimental results.} is given that contains 140 points in two classes (70 data points in each class) in which each instance $ \vec{x}_i $ is 4 dimensional. 

\begin{figure}
	\centering
	\begin{tabular}{c}
	\includegraphics[width=0.49\textwidth]{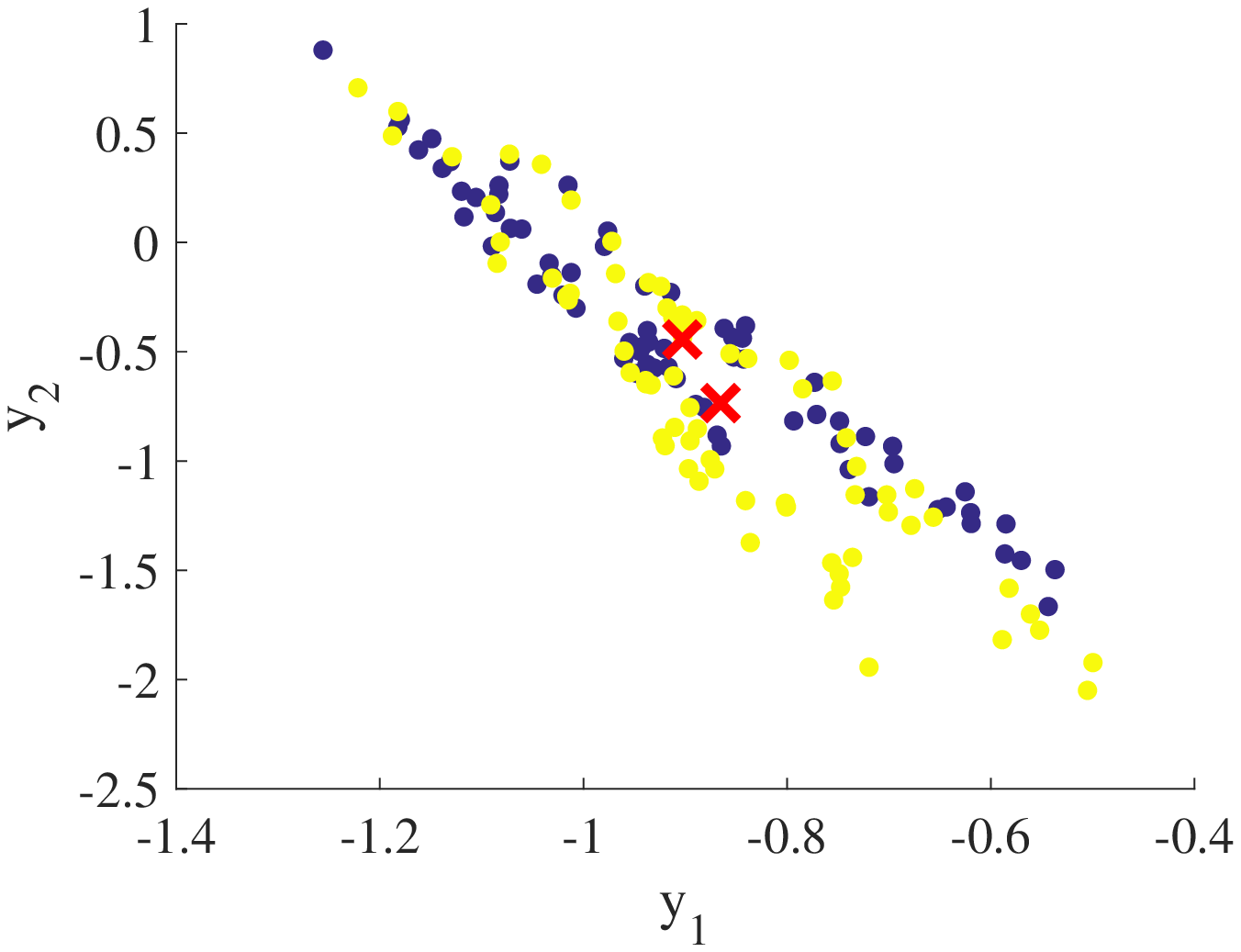} \\
	(a)\\
	\includegraphics[width=0.49\textwidth]{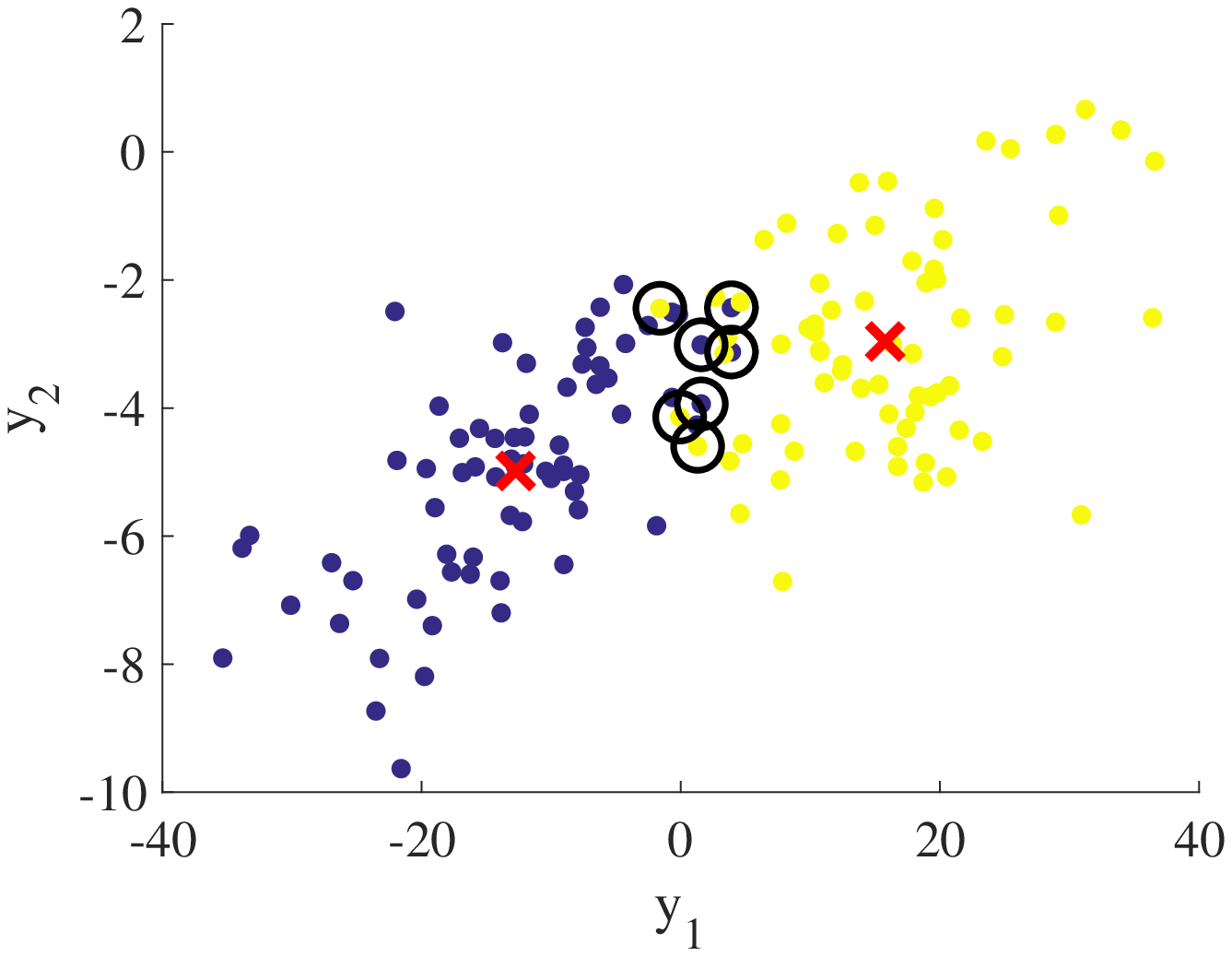}\\
	(b)
	\end{tabular}
	\caption{Application of ODD to the Crab gender dataset (4 dimensions) (a) before, (b) after, optimization of the parameters of $ F $.}
	\label{fig:proposed-transformed-Example}
\end{figure}

We set the transformation $ F $ as $ F(\vec{x}')=\vec{x}'M_{5 \times 2} $, where $ \vec{x'}_i=<x_i^1,x_i^2,x_i^3,x_i^4,1> $, and $ \vec{x}^k_i $ is the $ k^{th} $ element of the instance $ \vec{x}_i $. We then solve the optimization problem, $ \Omega $ (Eq. \ref{eq:proposed-obj-single}) to find the matrix $ M $. After optimization, we found 
\[M=\begin{bmatrix}
	9.169 & -99.006 & 80.214 & -6.983 & -12.605\\ 
	7.111 & -11.865 & -36.473 & 39.129 & -4.548
\end{bmatrix}^T
\] where $ T $ is the transpose operator. The transformed dataset before and after optimization of $ M $ has been shown in Fig. \ref{fig:proposed-transformed-Example}.a and b (filled circles), respectively. The figures also indicate the center of the distributions of transformed instances (crosses) as well as misclassified instances when the value of $ f $ was thresholded by an arbitrary value 0.5. With this threshold, the algorithm has classified 133 instances (over all 140) correctly. With an optimized threshold, this could be improved to 137 correctly classified cases.

\ignore{
\subsection{Extension to multiple layers}
\subsection{Linear separable dataset}
We will prove that ODD with linear transformation (i.e., F=...) function can separate any linearly separable datasets.

Assume two arbitrary sets of points that are linearly separable by a hyperplane H in a n dimensional space. There exist a linear transformation $ T_{n \times n} $ that transforms n-1 coordinate axes on H, the nth axis will be the norm of H. In the new coordinate axes the 

Assume that the are $ m_1 $ 
}

\subsection{Comparison with LDA}
\label{sec:comparewithLDA}
In this section we compare ODD with Direct-LDA on some artificially-generated datasets to demonstrate the differences between the methods.
\begin{itemize}
	\item \textbf{Db1}: contains two sets of time series with identical length (300 samples). Both series are in the form of $ a \sin(2\pi zt)+10r\cos(20\pi t)+10r\cos(400\pi t) + N(0,30) $, where $ N(0, 30) $ generates random values from a normal distribution with mean equal to zero and standard deviation 30. The value of $ a $ is picked uniformly randomly from $ [15,25] $ for both series. The frequency $ z $ is picked randomly (uniform distribution) from $ [65, 75] $ for 50 instances while it is picked randomly (uniform distribution) from $ [15, 25] $ for 500 instances. The first 50 instances are labeled as 1 and the rest are labeled as 2. A similar set is used for testing purposes but with 500 instances from each class.
	\item \textbf{Db2}: similar to the first dataset, but this time we have 4 different frequency ranges: $ [10, 20], [30, 40], [50, 60], [70, 80] $ in 4 classes. We place 50, 250, 500, and 10 instances from each time series for training and the same number of instances for testing. 
	\item \textbf{Db3}: two sets of time series, 100 instances in the first and 1000 in the second. Each sample of each series is generated by $ N(a,b) $. For the time series in the first group it is either $ a=40 $ and $ b\in[95, 105] $ (uniform distribution) OR $ b=80 $ and $ a\in[49,51] $ (uniform distribution). For the time series in the second group it is either $ a=40 $ and $ b\in[55, 65] $ (uniform distribution) OR b=80 and $ a\in[29,31] $ (uniform distribution). It is clear that these two sets are not distinguishable by their variance or mean, but both at the same time.
	\item \textbf{Db4}: was taken from "Figure 3" of \cite{gionis2007clustering}. It includes 788 instances in 7 classes in 2 dimensions\footnote{The data is available online at https://cs.joensuu.fi/sipu/datasets/.}. 
\end{itemize}

\begin{figure}
		\includegraphics[width=0.49\textwidth]{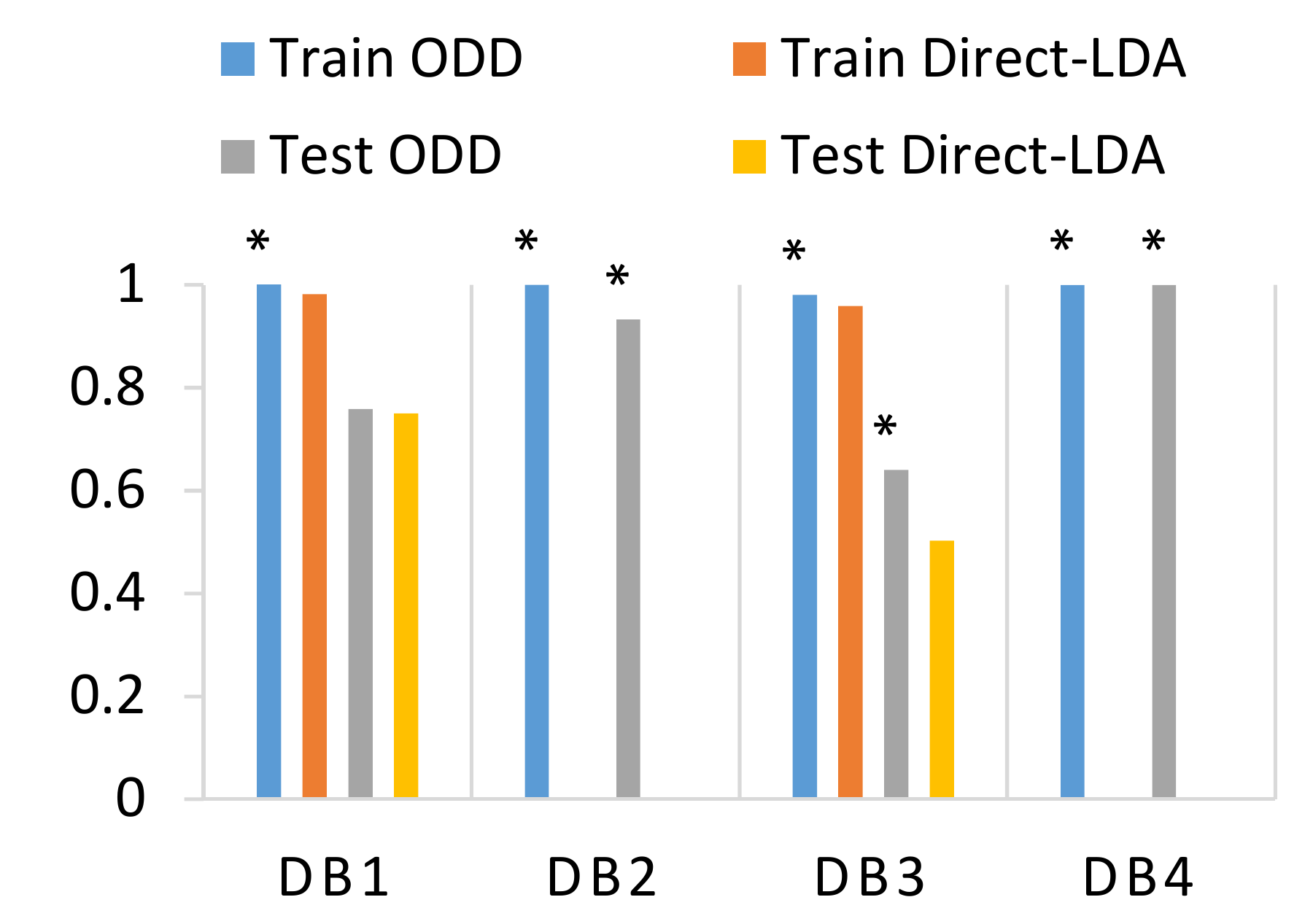}\\
	\caption{The results of the accuracy (area under the curve) of Direct-LDA in comparison to ODD. Star indicates the significance of the comparison (0.05 confidence, t-test). Direct-LDA could not find any solution for Db2 and Db4.}
	\label{fig:LDA-ODD}
\end{figure}

We applied ODD and Direct-LDA to these four datasets, results have been reported in Fig. \ref{fig:LDA-ODD}. For both methods we used the distance between the instances and the closest distribution center without any thresholding (function $f$, see section \ref{sec:discriminatorf}). Clearly, ODD outperforms Direct-LDA in all training sets. The most important denominator in all of these sets was that either the number of dimensions was smaller than the number of classes (e.g., Db4) or the number of instances in at least one of the classes was smaller than the number of dimensions (e.g., Db3). Both of these scenarios cause Direct-LDA to fail (the method could not find any solution for Db2 and Db4) while ODD performs fine in these scenarios. Note that the second scenario is very common in time series, i.e., large number of samples, each represent one dimension, and small number of instances for each class.

\section{Experimental results}
\label{sec:experiment}
In this section we provide comparative results of the algorithm with other state-of-the-art classification methods. We apply ODD together with other methods to 12 standard classification problems.

\subsection{Comparisons procedure}
In this subsection we introduce the datasets, pre-processes, and algorithms settings used for comparisons.
\subsubsection{Datasets}
We use 12 datasets for comparison among different classifiers, namely, Breast cancer (BC), Crab gender (GD), Glass chemical (GC), Parkinson (PR) \cite{parkinsonDb}, Seizure detection (SD) \cite{temko2015detection}, Iris (IR), Italian wines (IW), Thyroid function (TF), Yeast dataset (YD) \cite{horton1996probabilistic}, Red wine quality (RQ) \cite{redwinequalityDb}, White wine quality (WQ) \cite{redwinequalityDb}, and Handwritten digits (HD). Main characteristics of these datasets have been reported in Table \ref{Tab:dbs}. These datasets have been used frequently in previous classification studies as standard benchmarks. We used the one-vs-rest presentation for the classes, hence, the class of each instance was represented by a binary vector with the length $ c $.

\begin{table}[]
	\centering
	\caption{The datasets used for comparison purposes in this paper. $ n $ is the number of variables and $ c $ is the number of classes in each dataset. The number of instances in each class has been reported in the last column.}
	\label{Tab:dbs}
	\begin{tabular}{|l|l|l|l|}
		\hline
		\textbf{Dataset name} & \textbf{$ n $} & \textbf{$ c $} & \begin{tabular}[c]{@{}l@{}}\textbf{\# instance}\\ \textbf{in each class}\end{tabular}\\ \hline
		BC & 9 & 2 & $ <458,241> $ \\ \hline
		CG & 6 & 2 & $ <100,100> $ \\ \hline
		GC & 9 & 2 & $ <51,163> $ \\ \hline
		PR & 22 & 2 & $ <48,147> $ \\ \hline
		SD* & ? & 2 & ? \\ \hline
		IR & 4 & 3 & $ <50,50,50> $ \\ \hline
		IW & 13 & 3 & $ <59,71,48> $ \\ \hline	
		TF & 21 & 3 & $ <166,368,6666> $ \\ \hline
		YD & 8 & 10 & \begin{tabular}[c]{c} $ <463,5,35,44,$ \\ $51,163,244,$ \\ $429,20,30> $ \end{tabular} \\ \hline
		RQ & 11 & 6 & \begin{tabular}[c]{c} $ <10,53,681$ \\ $,638,199,18> $ \end{tabular}\\ \hline
		WQ & 11 & 7 & \begin{tabular}[c]{c} $ <20,163,1457,$ \\ $2198,880,175,5> $ \end{tabular}\\ \hline
		HD & 784 & 10 & \begin{tabular}[c]{c} $ <6903,7877,6990,$ \\ $7141,6824,6313 ,$ \\ $6876,7293,6825,6958> $ \end{tabular} \\ \hline
	\end{tabular}
	\vspace{1ex}
	
	\raggedright *The seizure detection dataset includes 12 patients, each of them has their own number of variables and instances in different classes. See Table \ref{Tab:SDDetails}.
\end{table}

The SD dataset includes interacornial electroencephalogram (iEEG) of 12 patients (4 dogs and 8 Human) with variable number of channels (Table \ref{Tab:SDDetails} shows the details of this dataset)\cite{temko2015detection}. There are 2 classes, namely seizure (ictal) and no-seizure (interictal), in the dataset with various number of instances and iEEG channels for each patient. While each seizure event might take up to 60 seconds, each instance of ictal or interictal in the dataset indicates only 1 second of an event from all iEEG channels. As the properties of the signals that belong to the same ictal event could be similar, including different segments of a single event in both training and testing sets may simplify the problem. Hence, we used all ictal segments that belonged to the same seizure event in either testing or training set, but not both. The segment index is available with the dataset that could be used to reconstruct signals from the same event. This procedure is usually used for cross-validation in seizure detection and prediction literature \cite{temko2015detection}. 

\begin{table}[]
	\centering
	\caption{Details of the seizure detection (SD) dataset. The first value in the last column is the number of instances of seizure (one second each) and the second number is the number of instances of non-seizure (one second each). The seizure instances are one-second segments from different seizure events.}
	\label{Tab:SDDetails}
	\begin{tabular}{|l|l|l|l|}
		\hline
		\textbf{Patient} & \textbf{\# of Channels} & \begin{tabular}{c} \textbf{Sampling}\\\textbf{ rate (Hz)} 
		\end{tabular}& \begin{tabular}[c]{@{}l@{}}\textbf{\# instance}\\ \textbf{in each class}\end{tabular}\\ \hline
		Subject 1 & 16 & 400 & $ <178,418> $ \\ \hline
		Subject 2 & 16 & 400 & $ <172,1148> $ \\ \hline
		Subject 3 & 16 & 400 &  $ <480,4760> $ \\ \hline
		Subject 4 & 16 & 400 &  $ <257,2790> $ \\ \hline
		Subject 5 & 68 & 500 &  $ <70,104> $ \\ \hline
		Subject 6 & 16 & 5000 &  $ <151,2990> $ \\ \hline
		Subject 7 & 55 & 5000 &  $ <327,714> $ \\ \hline
		Subject 8 & 72 & 5000 &  $ <20,190> $ \\ \hline
		Subject 9 & 64 & 5000 &  $ <135,2610> $ \\ \hline
		Subject 10 & 30 & 5000 &  $ <225,2772> $ \\ \hline
		Subject 11 & 36 & 5000 &  $ <282,3239> $ \\ \hline
		Subject 12 & 16 & 5000 &  $ <180,1710> $ \\ \hline
	\end{tabular}
\end{table}

\subsubsection{Pre-processes and performance measure}
We preprocessed the instances in the SD dataset by calculating the fast Fourier transform of each channel and concatenating these transformed signals to generate one large signal (FFT of the channels one after another). The length of this signal is a function of the number of channels of the iEEG device. We used frequencies from 1 to 50 Hz only, as it showed sufficiently accurate presentation of seizure \cite{temko2015detection}. For subject 1, for example, the preprocessed signal was $ 16 \times 49 $ samples long (16 channels, 1 Hz to 50 Hz FFT). The classifiers were trained for each patient independently. The training set was generated by selecting 70\% of the seizure events randomly and the rest of the seizure events were left for test. We also selected 70\% of non-seizure signals randomly for training and left the rest for test. For each selected set of training signals we trained KNN, MLP, SVM, DTR, LDA, and ODD models and tested their performances on the remaining instances. This was done for 50 independent runs to decrease the impact of possible biased selection in train and test sets. As the number of instances in each class for SD dataset is imbalance (the ratio ictal to interictal is almost 2:19 in average), this dataset forms a good test for the models. 

For other datasets, the variables were normalized to $ [-1,1] $ for the instances in the train set. The same mapping that normalized the training set was then applied to the test set to ensure both train and test sets are in the same domain. The variables with the variance equal to zero across the whole dataset were completely removed from further calculations. For HD dataset, we also used principle component analysis \cite{hotelling1933analysis} to reduce the number of dimensions to 50, prior to normalization. 

For each dataset (except for HD), we randomly selected 70\% of instances in each class for training purposes and left the rest for tests. After this selection, all methods were applied to the training set and then their performances on both training set and test set were measured. The selection and modeling were performed for 50 independent runs to ensure the results are not biased towards a specific combination of the instances in the train set. For each tune, the training set was remained unchanged for all methods. For HD dataset (70,000 instances), the number of selected instances for training purposes at each run was set to 5\% (rather than 70\%) from each class and the remaining instances were left for testing purposes. 

We used area under the curve (AUC) of the recipient operation curve (ROC) \cite{hanley1982meaning} as a performance measure. For multiclass problems, the overall performance was calculated by averaging AUC for each class separately. As MLP and ODD are generative classifiers, we used the best threshold found on the ROC curve of the training set to threshold the results of the test set and convert the results to discriminative. The same threshold was used for testing purposes.

We used the t-test (confidence 0.05) for statistical comparisons between the performance of each method and ODD.  

\subsubsection{Algorithms settings}
We compare the results of ODD with NBY, SVM (we used one-vs-all strategy to enable SVM to deal with multiple classes), MLP (with the number of neurons in the hidden layer equal to $ p $ (the dimension of the transformation $F$ in ODD) with Levenberg-Marquardt for backpropagation of weights errors), LDA (we used one-vs-all strategy to enable LDA to deal with multiple classes), DTR, and KNN (5 neighbors model). We used Matlab 2016b for implementations and tests. For SVM and LDA we used linear kernels. We tested three settings for ODD that are $ ODD_1 $ for which $ p=1 $ and $ F(\vec{x})=\vec{x}M_{n' \times p} $, $ ODD_l $ for which $ p=c $ and $ F(\vec{x})=\vec{x}M_{n' \times p} $, and $ ODD_n $ for which $ p=c $ and $ F(\vec{x})=tanh(\vec{x}M_{n' \times p}) $. Note that all derivatives of ODD could handle multiclass classification with no need for external strategies.

Stopping criteria for MLP was set to gradient $ < $ 1e-8 or 1000 iterations. For $ ODD_1 $ and $ ODD_l $, the stopping criteria was 100 iterations (constant) of CMAES and then 100 iterations of QN to ensure efficient exploration and exploitation. For CMAES, if the performance was not improving then the algorithm was terminated (improvement for the last 20 iterations was smaller than 0.0001). For QN, if gradient was smaller than 1e-8 then the algorithm was terminated. For large problems in these tests (HD and SD), we only used ES for 500 iterations. 

\subsection{Comparison with existing classifiers}
\label{sec:experimentOverall}
Table \ref{Tbl:resultsCompactAllDiscr} shows comparative results\footnote{Details of this experiment is available in Appendix A.} of tested methods for all datasets except for SD. The value in the row $ i $ column $ j $ shows $ P_{i,j}-G_{i,j} $ where $ P_{i,j} $ is the number of datasets for which the ODD type indicated in column $ j $ performs significantly (based on t-test, confidence 0.05) better than the method indicated in the row $ i $ and $ G_{i,j} $ is the number of datasets for which the method indicated in the row $ i $ performs significantly better than the ODD type indicated in column $ j $, for the performance measure indicated in the "Measure" column. For example, the value 3 in the row 3 (LDA), column 2 ($ODD_n$) for the measure "Test" indicates that the number of datasets for which $ODD_n$ performs significantly better than LDA is 3 datasets (over 11) more than number of datasets for which LDA performs significantly better than $ODD_n$.

Clearly the running time of all derivatives of $ ODD $ is significantly longer than the running time of NBY, SVM, LDA, DTR, and KNN. In comparison to MLP, however, $ ODD_l $ requires significantly less time in majority of datasets. This was not as good for $ ODD_n $ mainly because of the overhead of the non-linear function. For $ ODD_1 $, this running time was significantly less than MLP's in majority of  datasets. 

The performance of $ ODD_l $ was significantly better than all other methods in majority of datasets in the test sets (all values in that column are positive). The algorithm also performed significantly better than other methods in majority of datasets in the training set except in comparison to MLP and DTR. 

The performance of $ ODD_1 $ was significantly better than all other methods in majority of datasets for test sets except in comparison to MLP and LDA. For training sets, $ ODD_1 $ performs significantly better than KNN, LDA, SVM, and NBY in majority of datasets while DTR and MLP perform better than $ ODD_1 $ in majority of datasets for training set. 

The performance of $ ODD_n $ was significantly better than all other methods in majority of datasets in the test sets except in comparison to MLP, where there is a draw (for 5 dataset MLP performs better while for 5 $ ODD_n $ performs better). $ ODD_n $ also performed significantly better than other methods in the training set in majority of datasets except in comparison to MLP and DTR. 

\begin{table}[]
	\centering
	\caption{Comparison results among different classification algorithms. Each row indicates the comparative results with one type of ODD in terms of time, training set performance, and test set performance. Positive values indicate better performance (in terms of different measures) of the ODD types.}
	\label{Tbl:resultsCompactAllDiscr}
	\begin{tabular}{|l|l|l|l|l|l|l|l|l|l|}
		\hline
		  & \multicolumn{3}{l|}{$ ODD_l $} & \multicolumn{3}{l|}{$ ODD_n $} & \multicolumn{3}{l|}{$ ODD_1 $} \\ \hline
		Measure &  \rotatebox[origin=c]{270}{Time}   &   \rotatebox[origin=c]{270}{Train}   &  \rotatebox[origin=c]{270}{Test}   &   \rotatebox[origin=c]{270}{Time}   &   \rotatebox[origin=c]{270}{Train}   &  \rotatebox[origin=c]{270}{Test}  &   \rotatebox[origin=c]{270}{Time}   &   \rotatebox[origin=c]{270}{Train}   &  \rotatebox[origin=c]{270}{Test} \\ \hline
		{\textbf{NBY}}  & -11     & 11       & 7      & -11      & 10       & 4      & -11     & 7        & 5      \\ \cline{2-10} \hline
		{\textbf{SVM}}  & -11     & 11       & 8      & -11      & 11       & 7      & -11     & 7        & 5      \\ \cline{2-10} \hline
		{\textbf{LDA}}  & -11     & 9        & 2      & -11      & 9        & 3      & -11     & 5        & -1     \\ \cline{2-10} \hline
		{\textbf{MLP}}  & 7       & -11      & 1      & 1        & -11      & 0      & 9       & -11      & -2     \\ \cline{2-10} \hline
		{\textbf{DTR}}  & -11     & -5       & 7      & -11      & -4       & 7      & -11     & -7       & 5      \\ \cline{2-10} \hline
		{\textbf{KNN}}  & -11     & 6        & 3      & -11      & 7        & 3      & -11     & 3        & 1      \\ \cline{2-10} \hline
	\end{tabular}
\end{table}

\subsection{SD dataset results}
Table \ref{Tbl:resultsCompactSD}\footnote{Details of this experiments are available in Appendix B} shows comparative results of tested methods for SD dataset. $ ODD_l $ and $ ODD_1 $ performed significantly better than all other methods in majority of subjects for the test set. For the training set, however, $ ODD_l $ and $ ODD_1 $ performed better than all methods except LDA and SVM for majority of subjects. $ ODD_n $ performed better than all other methods for training set for majority of subjects except in comparison to SVM, LDA, and MLP. For test sets, the algorithm performs significantly better than all other methods in majority of subjects except in comparison to LDA.

\begin{table}[]
	\centering
	\caption{Comparative results among different classification algorithms when they were applied to SD dataset. The values in the table are similar to what was discussed for Table \ref{Tbl:resultsCompactAllDiscr}.}
	\label{Tbl:resultsCompactSD}
	\begin{tabular}{|l|l|l|l|l|l|l|l|l|l|}
		\hline
		& \multicolumn{3}{l|}{$ ODD_l $} & \multicolumn{3}{l|}{$ ODD_n $} & \multicolumn{3}{l|}{$ ODD_1 $} \\ \hline
		Measure &  \rotatebox[origin=c]{270}{Time}   &   \rotatebox[origin=c]{270}{Train}   &  \rotatebox[origin=c]{270}{Test}   &   \rotatebox[origin=c]{270}{Time}   &   \rotatebox[origin=c]{270}{Train}   &  \rotatebox[origin=c]{270}{Test}  &   \rotatebox[origin=c]{270}{Time}   &   \rotatebox[origin=c]{270}{Train}   &  \rotatebox[origin=c]{270}{Test} \\ \hline
		{\textbf{NBY}}  & -12     & 12       & 11     & -12      & 12       & 10     & -12     & 12       & 11     \\ \cline{2-10} \hline
		{\textbf{SVM}}  & -12     & -6       & 12     & -12      & -9       & 12     & -12     & -6       & 12     \\ \cline{2-10} \hline
		{\textbf{LDA}}  & -12     & -1       & 6      & -12      & -5       & -1     & -12     & 0        & 5      \\ \cline{2-10} \hline
		{\textbf{MLP}}  & -12     & 3        & 12     & -12      & -1       & 11     & -12     & 3        & 12     \\ \cline{2-10} \hline
		{\textbf{DTR}}  & -12     & 10       & 10     & -12      & 6        & 9      & -12     & 10       & 10     \\ \cline{2-10} \hline
		{\textbf{KNN}}  & -12     & 12       & 12     & -12      & 12       & 10     & -12     & 12       & 12     \\ \cline{2-10} \hline
	\end{tabular}
\end{table}

In terms of running time, all types of ODD performed significantly slower than other method.

\subsection{Sensitivity to between-class imbalance datasets}
ODD is not sensitive to the number of instances in each class. The reason is that ODD optimizes characteristics (mean and spread) of a class that is independent of the number of instances in that class. It is easy to see that, assuming instances in each class represent the exact mean and spread of that class, the performance of ODD is independent of the number of instances in the classes. Although in reality the instances in each class might not reflect the exact mean and spread of the class, it is expected that these parameters are preserved in the given data to some extent to enable the algorithm to perform well. Similar assumption is required by any other classification algorithm (i.e., training set must reflect the characteristics of the data in each class), otherwise, it is impossible to design any effective classifier. 

\begin{figure}
	\centering
	\includegraphics[width=0.49\textwidth]{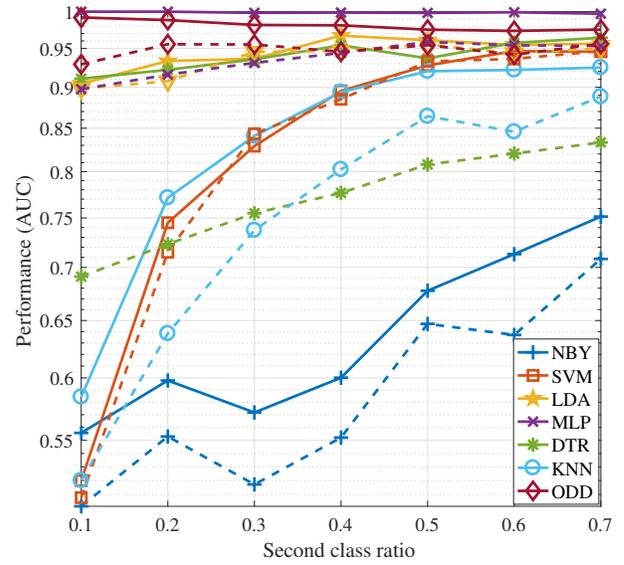} 
	\caption{Comparative results in terms of the sensitivity to the number of instances in each class. The dashed lines are the test results while the solid lines are the training results. ODD and MLP are affected only slightly by imbalanced between-class number of instances. The CG dataset was used for this test.}
	\label{fig:proposed-imbalancesensitivy}
\end{figure}

We test the between-class imbalance sensitivity of ODD by using the CG dataset. We divide this dataset to two subsets, one for training and the other for test, with 70\% of instances from the first class were always present in the train set. The ratio of the instances from the second class was $ 100r\% $, $ r \in \{0.1, 0.2, ..., 0.7\} $. As the number of instances in each class is equal to 100 in this dataset, the changes in $r$ causes changes in the balance of the number of instances in the training set. We run NBY, SVM, LDA, DCT, KNN, and ODD for training and predict the class of the remaining instances. The average of the performance (area under the curve, AUC, was used for performance measure) of the methods over 50 runs has been reported in Fig. \ref{fig:proposed-imbalancesensitivy}. For ODD, $ p $ was set to 1 to ensure a fair comparison with SVM, hence, $ F(\vec{x})=\vec{x}M_{n' \times 1} $ where $ n'=7 $ as the number of variables in the dataset is 6. 

Figure \ref{fig:proposed-imbalancesensitivy} that the performance of ODD and MLP changed slightly when the class ratio was changed from 0.1 to 0.7. This, however, is not correct for SVM, NBY, LDA, DCT, and KNN and these methods have been affected significantly by changing the imbalance ratio.

\subsection{Overfitting}
Over-fitting is a very common issue among classification methods. It is usually measured by the extent the performance of a classification method could be generalized to unseen instances. To formulate overfitting for a classification method, we calculate the performance extension index defined by: 
\begin{equation}
G_{indx}=\frac{\text{Test performance}}{\text{Train performance}}\times\text{Test performance}
\end{equation}
This index indicates to what extent the performance of the classifier on the train data could be generalized to the unseen data (the fraction) while also takes into account how well the algorithm performs in the testing set (the second term). The second term has been used in this formula to penalize algorithms that have a very low performance (e.g., 50 percent accuracy in both test and train) on both training set while their bad performance is extendable.

\begin{figure}
	\centering
     \includegraphics[width=0.49\textwidth]{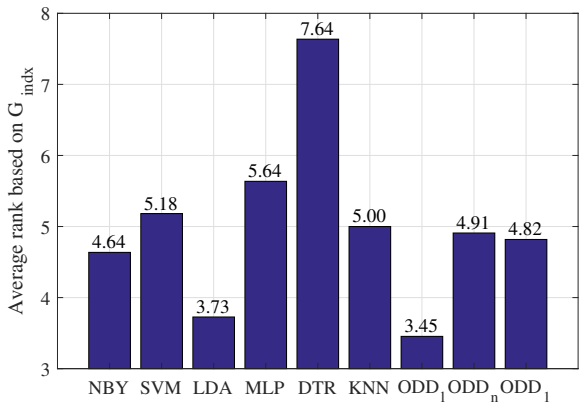}     
	\caption{Comparison results among different types of ODD and other classifiers in terms of $G_{indx}$. The vertical axis indicates the average rank of the methods in terms of their $G_{indx}$. The smaller the rank the more successful the algorithm was in generalization of its performance according to the $G_{indx}$ measure.}
	\label{fig:canc-gind}
\end{figure}

The larger the value of $G_{indx}$ is, the better the algorithm can generalize its performance to the training set, taking into account how good is the algorithm performance in the test set. We calculated the $G_{indx}$ for all methods based on their results in section \ref{sec:experimentOverall}. Figure \ref{fig:canc-gind} shows the average ranking of the algorithms based on their $G_{indx}$ value over all 11 datasets (the smaller the ranking the better). The figure indicates that $ODD_l$ has the best ranking among these methods in terms of its $G_{indx}$. LDA is in the second place with a small margin and NBY is in the third place. Also, $ODD_n$ has better generalization ability comparing to all non-linear classification methods tested in this article (i.e., MLP, DTR, and KNN). 

\section{Conclusion and future works}
\label{sec:conclusions}
This paper introduced a new method based on optimization of distribution differences (ODD) for classification. The aim of the algorithm was to find a transformation to minimize the distance between instances in the same class while maximize the distance between gravity centers of the classes. The definitions of ODD allows the use of any transformation, however, this paper only tested linear transformation and a particular non-linear transformation. The algorithm was applied to 12 benchmark classification problems and results were compared with state-of-the-art classification algorithms. Comparisons showed that the proposed algorithm outperforms previous methods in most datasets. Results also showed that the method is less sensitive to the between class imbalance and has a better ability to maintain its performance for the instances that have not been used in the training process. 

This paper only experimented with reducing the number of dimensions to a value equal to the number of classes. However, ODD design allows flexible dimensionality reduction/increase that could be beneficial in some datasets. Hence, as a future work, we are going to design an adaptive method to increase the number of dimensions in the linear transformation ($ p $) gradually to find the best size for the transformation matrix. A small transformation matrix could be beneficial especially for hardware implementations. In addition, the method could be extended to make use of a multiple layers in which each layer reduces the number of dimensions that could improve the performance of the algorithm for some classification tasks. Another interesting direction would be to change the stopping criteria and test its impact on overfitting, i.e., stop the algorithm as soon as the instances were separable in different classes. 

\section*{Appendix-A}
Table \ref{Tbl:comparisonResultsDisc} shows the average results of 50 independent runs for all algorithms. The values in the table have been prefixed by three characters. The character in position one, two, and three after each value at a specific row and column indicate the results of the statistical test (t-test with confidence 0.05) between the method indicated in that column and $ ODD_l $, $ ODD_n $, and $ ODD_1 $, respectively, for the dataset at that row.  "*", "-", and "+" at a position indicate that that result is statistically worst, the same, or better than $ ODD_l $, $ ODD_n $, and $ ODD_1 $, depending on the position of the character. For example, the value "$96.37^{*-*}$" in the row "BC", measure "Test", column NBY indicates that the average performance (AUC) of the method NBY was $96.37$ on the training set, that was significantly worse than $ODD_l$ and $ODD_1$ while statistically the same comparing to $ODD_n$

\begin{table*}[]
	\centering
	\caption{Comparison results among 6 well-known classification methods with ODD. $ ODD_l $ refers to a setting where $ p=c $, $ c $ is the number of classes, and a linear kernel is used for $ F $. $ ODD_n $ uses the same value for $ p $ with $ F(\vec{x})=tanh(\vec{x}M_{n' \times p}) $. $ ODD_1 $ uses a linear $ F $ while sets $ p=1 $. The values in each row is the average (over 50 runs) of results. The row "Time" reports the average of time in milliseconds over 50 runs.}
	\label{Tbl:comparisonResultsDisc}
	\begin{tabular}{|l|l|l|l|l|l|l|l|l|l|l|}
		\hline
		Dataset & Measure & \textbf{NBY} & \textbf{SVM} & \textbf{LDA} & \textbf{MLP} & \textbf{DTR} & \textbf{KNN} & \textbf{$ ODD_l $} & \textbf{$ ODD_n $} & \textbf{$ ODD_1 $} \\ \hline
		\multirow{4}{*}{BC}  & Time & 20.4 & 12.6 & 10.5 & 991.9 & 8.8 & 8.4 & 99 & 423.7 & 63.2  \\ \cline{2-11} 
& Train & 96.37*** & 96.9*** & 95.26*** & \textbf{98.63}+++ & 97.96+-+ & 97.55*** & 97.77 & 97.91 & 97.77  \\ \cline{2-11} 
& Test & 96.37*-* & 96.49*-* & 95.13*** & 95.42*** & 93.98*** & 96.71--- & \textbf{97.06} & 96.6 & 97.06  \\ \hline
\multirow{4}{*}{CG}  & Time & 6.7 & 5.4 & 5.3 & 154.3 & 5 & 4.6 & 148.2 & 343.6 & 98.2  \\ \cline{2-11} 
& Train & 76.39*** & 94.86*** & 95.79*** & \textbf{99.66}+++ & 96.16*** & 93.03*** & 96.7 & 97.46 & 96.93  \\ \cline{2-11} 
& Test & 69.93*** & 94.13--- & \textbf{94.73}+++ & 94.33--- & 82.07*** & 86.93*** & 94.08 & 94.07 & 94.25  \\ \hline
\multirow{4}{*}{GC}  & Time & 7.9 & 5.2 & 5.2 & 527.8 & 4.8 & 4.5 & 124.1 & 366.1 & 84.2  \\ \cline{2-11} 
& Train & 87.61*** & 89.86*** & 89.58*** & \textbf{98.85}+++ & 96.93+++ & 93.08*** & 95.35 & 96.07 & 95.36  \\ \cline{2-11} 
& Test & 86.58*** & 87.71*** & 85.7*** & 86.74*** & 88.73*** & 87.04*** & 90.44 & \textbf{90.63} & 90.21  \\ \hline
\multirow{4}{*}{PR}  & Time & 13.4 & 5.3 & 5.5 & 720.5 & 5.1 & 4.6 & 325.3 & 843 & 165.6  \\ \cline{2-11} 
& Train & 78.32*** & 79.42*** & 84.99*** & \textbf{97.52}+++ & 96+++ & 93.05+-+ & 88.03 & 93.86 & 87.6  \\ \cline{2-11} 
& Test & 76.64--- & 74.83-*- & 77.64--+ & 75.83-*- & 78.24--+ & \textbf{85.56}+++ & 76.29 & 78.35 & 75.12  \\ \hline
\multirow{4}{*}{IR}  & Time & 6.2 & 9.6 & 6 & 255.2 & 4.7 & 4.6 & 138.9 & 421.2 & 86.1  \\ \cline{2-11} 
& Train & 97.06*** & 80.7*** & 98.53*** & \textbf{99.68}+++ & 98.47*** & 97.64*** & 98.69 & 99.32 & 98.69  \\ \cline{2-11} 
& Test & 96.43--- & 78.73*** & \textbf{98.2}+++ & 95.7*** & 96.2--* & 97.03--- & 96.88 & 96.59 & 96.86  \\\hline
\multirow{4}{*}{IW}  & Time & 10.4 & 9.9 & 5.9 & 83.7 & 4.9 & 4.1 & 316.2 & 719.8 & 136.4  \\ \cline{2-11} 
& Train & 98.94**+ & 99.31**+ & 99.99+++ & \textbf{100}+++ & 98.85**+ & 98.26**+ & 99.78 & 99.76 & 97.33  \\ \cline{2-11} 
& Test & 97.94-++ & 98.04-++ & \textbf{98.88}+++ & 95.58**+ & 91.88*** & 97.48-++ & 97.56 & 96.83 & 93.83  \\ \hline
\multirow{4}{*}{TF}  & Time & 0 & 229.2 & 10.8 & 8431.5 & 8.4 & 5.2 & 685.1 & 4261.6 & 275.6  \\ \cline{2-11} 
& Train & N/A*** & 64.35*** & 62.57*** & 98.3+++ & \textbf{99.55}+++ & 70.03*** & 78.4 & 83.49 & 76.92  \\ \cline{2-11} 
& Test & N/A*** & 63.8*** & 61.99*** & 96.5+++ & \textbf{98.52}+++ & 64.83*** & 76.71 & 81.72 & 75.5  \\ \hline
\multirow{4}{*}{YD}  & Time & 0 & 100.7 & 18.3 & 15303.7 & 11.1 & 5.6 & 2601.5 & 6299.7 & 454.5  \\ \cline{2-11} 
& Train & N/A*** & 64.07*** & 76.74**+ & \textbf{85.26}+++ & 82.52+++ & 80.27--+ & 80.19 & 80.43 & 73.87  \\ \cline{2-11} 
& Test & N/A*** & 63.61*** & 76.23+-+ & \textbf{77.69}+++ & 67.1*** & 75.34-*+ & 75.26 & 75.92 & 69.71  \\ \hline
\multirow{4}{*}{RW}  & Time & 13.1 & 83.4 & 10.2 & 6335.4 & 13.5 & 4.3 & 1386.2 & 3871.9 & 395.9  \\ \cline{2-11} 
& Train & 64.82*** & 57.58*** & 62.13*** & 79.36+++ & \textbf{82.41}+++ & 66.3*** & 77.75 & 78.06 & 75.62  \\ \cline{2-11} 
& Test & 60.57*** & 57.37*** & 60.42*** & 70.67-+- & 60.16*** & 58.92*** & \textbf{70.71} & 68.68 & 69.92  \\ \hline
\multirow{4}{*}{WW}  & Time & 19 & 673.6 & 13.5 & 21432.9 & 35.9 & 4.8 & 1743.5 & 7361.5 & 547.3  \\ \cline{2-11} 
& Train & 61.74*** & 52.08*** & 58.53*** & 75.6+++ & \textbf{80.14}+++ & 67.58*** & 71.8 & 70.67 & 68.85  \\ \cline{2-11} 
& Test & 59.33*** & 52.01*** & 57.66*** & \textbf{71.05}+++ & 60.78*** & 59.18*** & 65.8 & 61.99 & 65.26  \\ \hline
\multirow{4}{*}{HD}  & Time & 92 & 2099.2 & 33.1 & 3222.5 & 127.9 & 7.7 & 20570.5 & 28141.2 & 941.2  \\ \cline{2-11} 
& Train & 93.1*-+ & 90.63**+ & 92.73**+ & 96.39+++ & 95.95+++ & \textbf{96.89}+++ & 93.26 & 93.13 & 78.08  \\ \cline{2-11} 
& Test & 92.34-++ & 89.23**+ & 92.06**+ & 92.46-++ & 85.11**+ & \textbf{95.1}+++ & 92.33 & 92.19 & 77.04  \\ \hline
	\end{tabular}
\end{table*}

\section*{Appendix-B}
Table \ref{Tbl:comparisonResultsSeizure} shows the average results of 50 independent runs for all algorithms when they were applied to SD. The values in the table have been postfixed by three characters that have the same definitions as in Table \ref{Tbl:comparisonResultsDisc}.

\begin{table*}[]
	\centering
	\caption{Results for seizure detection dataset.}
	\label{Tbl:comparisonResultsSeizure}
	\begin{tabular}{|l|l|l|l|l|l|l|l|l|l|l|}
		\hline
		Dataset & Measure & \textbf{NBY} & \textbf{SVM} & \textbf{MLP} & \textbf{LDA} & \textbf{DTR} & \textbf{KNN} & \textbf{$ ODD_l $} & \textbf{$ ODD_n $} & \textbf{$ ODD_1 $} \\ \hline
		\multirow{4}{*}{Subject 1}  & Time & 371.2 & 24.1 & 1924.4 & 94.8 & 68.1 & 7.6 & 6730 & 7652.3 & 3354.1  \\ \cline{2-11} 
		& Train & 83.73*** & \textbf{100}+++ & 100+++ & 100+++ & 99.72*** & 97.35*** & 100 & 99.83 & 100  \\ \cline{2-11} 
		& Test & 87.11*** & 95.15*** & 96.72*-* & 82.81*** & 89.77*** & 90.72*** & 97.2 & 96.76 & \textbf{97.23}  \\ \hline
		\multirow{4}{*}{Subject 2}  & Time & 410.5 & 48.2 & 3087.3 & 170.1 & 302.2 & 5 & 10848.9 & 13009.4 & 6720.9  \\ \cline{2-11} 
		& Train & 90.58*** & \textbf{100}--- & 100--- & 100--- & 99*** & 93.6*** & 100 & 100 & 100  \\ \cline{2-11} 
		& Test & 69.65*** & 79.44*** & \textbf{94.32}+++ & 77.54*** & 72.36*** & 87.98*** & 92.1 & 91.78 & 91.93  \\ \hline 
		\multirow{4}{*}{Subject 3}  & Time & 827.9 & 791.2 & 8542.1 & 452.5 & 2258.6 & 7.5 & 47136.3 & 52129.2 & 33319.5  \\ \cline{2-11} 
		& Train & 91.53*** & \textbf{100}+++ & 99.88++- & 96.47*** & 98.78*** & 96.54*** & 99.81 & 99.38 & 99.83  \\ \cline{2-11} 
		& Test & 91.07*** & 89.44*** & 95.3*-* & 91*** & 89.06*** & 89.43*** & \textbf{96.77} & 95.54 & 96.76  \\ \hline 
		\multirow{4}{*}{Subject 4}  & Time & 545.4 & 689.2 & 5606.9 & 292.3 & 1198.8 & 5.9 & 24416.3 & 28656.7 & 18166.7  \\ \cline{2-11} 
		& Train & 67.41*** & \textbf{100}+++ & 100+++ & 95.35*** & 98.53*** & 96.09*** & 99.79 & 99.65 & 99.81  \\ \cline{2-11} 
		& Test & 83.5*** & 77.13*** & 73.07*** & 80.52*** & 76.39*** & 82.61*** & \textbf{95.33} & 91.32 & 94.47  \\ \hline
		\multirow{4}{*}{Subject 5}  & Time & 1532.9 & 19.6 & 2163.2 & 102.5 & 53.4 & 4.7 & 9025.9 & 9108.6 & 5912.7  \\ \cline{2-11} 
		& Train & 93.56*** & \textbf{100}--- & 99.01--- & 100--- & 100--- & 98.32*** & 100 & 100 & 100  \\ \cline{2-11} 
		& Test & 86.85*** & 84.09*** & \textbf{96.52}--- & 79.54*** & 79.91*** & 87.01*** & 94.74 & 96.43 & 94.66  \\ \hline
		\multirow{4}{*}{Subject 6}  & Time & 636.6 & 192.5 & 5614.7 & 294.8 & 680.4 & 6.1 & 25360.2 & 29460.1 & 18817.2  \\ \cline{2-11} 
		& Train & 92.65*** & \textbf{100}+++ & 99.81*+* & 97.9*** & 97.65*** & 94.34*** & 99.97 & 99.6 & 99.97  \\ \cline{2-11} 
		& Test & 94.9*** & 95.03*** & 98.7*** & 96.7*** & 90.53*** & 91.02*** & 99.34 & 99.07 & \textbf{99.35}  \\ \hline
		\multirow{4}{*}{Subject 7}  & Time & 1413 & 388.7 & 5461.3 & 600.4 & 367.2 & 5.9 & 31242.9 & 32758.9 & 25726.8  \\ \cline{2-11} 
		& Train & 82.76*** & \textbf{100}-+- & 98.16--- & 100-+- & 99.71*-* & 90.92*** & 100 & 99.67 & 100  \\ \cline{2-11} 
		& Test & 60.84*** & 61.44*** & \textbf{73.55}+-+ & 60.87*** & 71.98+-+ & 55.91*** & 69.13 & 72.69 & 67.12  \\ \hline
		\multirow{4}{*}{Subject 8}  & Time & 1621.4 & 15.1 & 1619.4 & 127.2 & 42.9 & 4.7 & 11072.7 & 11317.2 & 7080.7  \\ \cline{2-11} 
		& Train & 99.47*** & \textbf{100}-+- & 99.04--- & 100-+- & 100-+- & 97.5*** & 100 & 99.86 & 100  \\ \cline{2-11} 
		& Test & 52*** & 53.75*** & 64.31*+* & 56.21*-* & 46.44*** & 57.78*+* & \textbf{71.49} & 55.47 & 71.36  \\ \hline
		\multirow{4}{*}{Subject 9}  & Time & 2234.3 & 1060 & 12547.6 & 7992.8 & 1991.4 & 8.9 & 97547 & 102059.8 & 76298.9  \\ \cline{2-11} 
		& Train & 80.29*** & \textbf{100}-+- & 99.98-+- & 100-+- & 99.96*+* & 94.44*** & 100 & 99.65 & 100  \\ \cline{2-11} 
		& Test & 91.81-+- & 85*** & 89.66*-* & 68.11*** & 77.67*** & 86.11*** & \textbf{91.84} & 89.34 & 91.79  \\ \hline
		\multirow{4}{*}{Subject 10}  & Time & 1038.9 & 335 & 7782.7 & 1103.1 & 573.7 & 7 & 51889.3 & 55736.9 & 37255.1  \\ \cline{2-11} 
		& Train & 96.21*** & \textbf{100}+-+ & 100+-+ & 100+-+ & 99.76*** & 90.89*** & 100 & 99.99 & 100  \\ \cline{2-11} 
		& Test & 85.87*** & 85.84*** & 93.31*+* & 86.96*** & 89.29*+* & 75*** & \textbf{95.55} & 88.52 & 95.4  \\ \hline
		\multirow{4}{*}{Subject 11}  & Time & 1592.4 & 808.9 & 9661.8 & 1988.7 & 2235.8 & 8.1 & 70730.1 & 74210.7 & 52923.8  \\ \cline{2-11} 
		& Train & 91.04*** & \textbf{100}-+- & 84.12*** & 99.98*+* & 99.42*** & 96.41*** & 100 & 99.96 & 100  \\ \cline{2-11} 
		& Test & 98.6*** & 99.03*** & 83.43*** & 98.38*** & 96.14*** & 99.21*** & \textbf{99.92} & 99.85 & 99.91  \\ \hline
		\multirow{4}{*}{Subject 12}  & Time & 447.5 & 111 & 3865.3 & 205 & 256.1 & 5.2 & 16811.3 & 19884.6 & 9818.7  \\ \cline{2-11} 
		& Train & 93.85*** & \textbf{100}+++ & 99.74*+* & 98.89*** & 98.33*** & 97.22*** & 99.98 & 99.05 & 99.98  \\ \cline{2-11} 
		& Test & 90.04*** & 91.77*** & \textbf{97.95}-++ & 88.63*** & 88.09*** & 93.06*** & 96.85 & 95.38 & 96.82  \\ \hline
	\end{tabular}
	\vspace{1ex}
	
	\raggedright Because of the size of these problems, we used ES with 500 iterations only.
\end{table*}

\small

\bibliographystyle{IEEEtran}
\bibliography{References}

\end{document}